\pdfoutput=1

\documentclass[11pt]{article}

\usepackage[]{acl}

\usepackage{flushend}

\usepackage{times}
\usepackage{latexsym}
\usepackage{booktabs}

\usepackage{amsmath}
\usepackage{amsfonts}
\usepackage{amssymb}
\usepackage{amsthm}
\usepackage{pgfplots} 
\usepackage{pgfplotstable} 
\usepackage{siunitx} 
\usepackage{graphicx} 
\usepackage{multirow}
\usepackage{subcaption}
\usepackage{dsfont}

\usepackage{hyperref}
\hypersetup{
    colorlinks=true,
    linkcolor=blue,
    filecolor=magenta,      
    urlcolor=cyan,
    pdftitle={Overleaf Example},
    pdfpagemode=FullScreen,
    }

\usepackage[T1]{fontenc}

\usepackage[utf8]{inputenc}

\usepackage{microtype}

\newtheorem{definition}{Definition}

\newcommand{\eos}{\left<\mathrm{eos}\right>}

\DeclareMathOperator*{\argmax}{arg\,max}

\usepackage[nameinlink]{cleveref}

\crefformat{equation}{Eq.~\textup{#2#1#3}}
\Crefformat{equation}{Eq.~\textup{#2#1#3}}

\crefformat{figure}{Figure \textup{#2#1#3}}
\Crefformat{figure}{Figure~\textup{#2#1#3}}

\crefformat{section}{Section \textup{#2#1#3}}
\Crefformat{section}{Section ~\textup{#2#1#3}}

\crefformat{table}{Table \textup{#2#1#3}}
\Crefformat{table}{Table ~\textup{#2#1#3}}

\title{Characterizing and addressing the issue of oversmoothing \\ in neural autoregressive sequence modeling}

\author{Ilia Kulikov\thanks{~~Equal contribution.} \\
  New York University \\
  \texttt{kulikov@cs.nyu.edu} \\\And
  Maksim Eremeev\footnotemark[1] \\
  New York University \\
  \texttt{eremeev@nyu.edu} \\\And
  Kyunghyun Cho \\
  New York University \\
  Genentech \\
  CIFAR Fellow in LMB}

\begin{document}
\maketitle
\begin{abstract}
Neural autoregressive sequence models smear the probability among many possible sequences including degenerate ones, such as empty or repetitive sequences.
In this work, we tackle one specific case where the model assigns a high probability to unreasonably short sequences. We define the oversmoothing rate to quantify this issue. After confirming the high degree of oversmoothing in neural machine translation, we propose to explicitly minimize the oversmoothing rate during training. We conduct a set of experiments to study the effect of the proposed regularization on both model distribution and decoding performance. We use a neural machine translation task as the testbed and consider three different datasets of varying size. Our experiments reveal three major findings. First, we can control the oversmoothing rate of the model by tuning the strength of the regularization. Second, by enhancing the oversmoothing loss contribution, the probability and the rank of $\eos$ token decrease heavily at positions where it is not supposed to be.
Third, the proposed regularization impacts the outcome of beam search especially when a large beam is used. 
The degradation of translation quality (measured in BLEU) with a large beam significantly lessens with lower oversmoothing rate, but the degradation compared to smaller beam sizes remains to exist. From these observations, we conclude that the high degree of oversmoothing is the main reason behind the degenerate case of overly probable short sequences in a neural autoregressive model.
\end{abstract}

\section{Introduction}
\label{sec:intro}

Neural autoregressive sequence modeling is a widely used scheme for conditional text generation. It is applied to many NLP tasks, including machine translation, language modeling, and conversation modeling \citep{cho2014learning, sutskever2014nmt,brown2020language,roller2021recipes}. Despite the substantial success, major issues still exist, and it is still an active area of research. Here we highlight two major issues which have been discussed extensively.

The first issue is the model assigning too high a probability to a sequence which is unreasonably shorter than a ground-truth sequence. \citet{stahlberg2019nmt} report evidence of an extreme case where the model frequently assigns the highest probability to an empty sequence given a source sequence in machine translation. In addition, \citet{koehn2017six} demonstrate that the length of generated translation gets shorter with better decoding (i.e., beam search with a larger beam.)

In the second issue, which is more often observed in open-ended sequence generation tasks, such as sequence completion, generated sequences often contain unreasonably many repetitions \citep{holtzman2019curious, Welleck2020Neural}. This phenomenon was partly explained in a recent year by \citet{welleck2020consistency}, as approximate decoding resulting in an infinitely long, zero-probability sequence.

In this work, we tackle the first issue where the model prefers overly short sequences compared to longer, often more correct ones. 
We assume that any prefix substring of a ground-truth sequence is an unreasonably short sequence and call such a prefix as a premature sequence. 
This definition allows us to calculate how often an unreasonably short sequence receives a higher probability than the original, full sequence does. This value quantifies the degree to which the probability mass is oversmoothed toward shorter sequences. We call this quantity an { \it oversmoothing rate}. We empirically verify that publicly available, well-trained translation models exhibit high oversmoothing rates.

We propose to minimize the oversmoothing rate during training together with the negative log-likelihood objective. 
Since the oversmoothing rate is difficult to minimize directly due to its construction as the average of indicator functions, we design its convex relaxation, to which we refer as an {\it oversmoothing loss}. This loss is easier to use with gradient-based learning.

We apply the proposed regularization to neural machine translation using IWSLT'17 and WMT tasks and observe promising findings. We effectively reduce the oversmoothing rate by minimizing the proposed oversmoothing loss across all tasks we consider.
We see the narrowing gap between the length distribution of generated sequences and that of the reference sequences, even when we increase the beam size, with a lower oversmoothing rate. 
Finally, by choosing the strength of the proposed regularization appropriately, we improve the translation quality when decoding with large beam sizes. We could not, however, observe a similar improvement with a small beam size.

\section{Background: Neural autoregressive sequence modeling}
\label{sec:background}

We study how a neural sequence model assigns too high probability to unreasonably short sequences due to its design and training objective. We do so in the context of machine translation in which the goal is to model a conditional distribution over a target language given a source sentence. More specifically, we consider a standard approach of autoregressive neural sequence modeling for this task of neural machine translation, where the conditional probability of a target sentence given a source sentence is written down as:\footnote{
In the rest of the paper, we often omit $X$ for brevity.
}
\begin{align}
\label{eq:chainrule}
	p(\mathbf{y}|\mathbf{x}) = \prod_{t=1}^{|\mathbf{y}|} p(y_t|y_{<t}, \mathbf{x}; \theta),
\end{align}
where $y_{<t}$ is a sequence of tokens up to (and not including) step $t$.
$\theta$ refers to the parameters of an underlying neural network that computes the conditional probability.
Each of the source and target sentences ends with a special $\left<\mathrm{eos}\right>$ token indicating the end of the sequence. As was demonstrated by \citet{newman2020eos}, this $\eos$ token is used by an autoregressive neural network to model the length of a sequence.

Given this parameterization, we assume a standard practice of maximum likelihood learning which estimates the parameters $\theta$ that maximizes the following objective function:
\begin{align*}
    L(\theta) &= \frac{1}{|D|} \sum_{n=1}^{N} \log p(\mathbf{y}^n|\mathbf{x}^n; \theta) + \mathcal{R}(\theta).
\end{align*}
$R$ is a regularization term that prevents overfitting, such as weight decay. 

Once training is done, we use this autoregressive model as a translation system by approximately solving the following optimization problem:
\begin{align*}
    \hat{\mathbf{y}}_\mathrm{map} = \argmax_{\mathbf{y}} p(\mathbf{y}|\mathbf{x}; \theta).
\end{align*}
We often resort to greedy decoding or beam search, both of which belong to a family of incomplete decoding algorithms~\citep{welleck2020consistency}. 

\section{Oversmoothing: the issue of premature sequences}
\label{sec:oversmoothing}

In this section, we carefully describe the issue of premature translation or premature sequence in autoregressive modeling, which has more often been referred to casually as the issue of oversmoothing in earlier studies \citep[see, e.g.,][]{shi2020neural}. To do so, we first define formally what we mean by a `premature sequence'. A premature sequence is a length-$t$ prefix of an original sequence, where $t$ is smaller than the length of the original sequence. In other words, length-$t$ prefix is defined as:
\begin{definition}[Length-$t$ prefix]
Given an original sequence $\mathbf{y}=(y_1,y_2,\ldots, y_T=\left<\mathrm{eos}\right>)$, the length-$t$ prefix is $\mathbf{y}_{\leq t}=(y_1, y_2, \ldots, y_{t-1}, \left<\mathrm{eos}\right>)$, where $1 \leq t < T$.
\end{definition}

With this definition, we make a reasonable assumption that most of such premature sequences are not valid sequences on their own. In the case of natural language processing, for instance, these premature sequences correspond to sentences that suddenly terminate in the middle. Only a few of these premature sequences may be a coherent, well-formed text.

A good autoregressive language model should then assign a lower probability to such an ill-formed premature sequence than that assigned to a well-formed original sequence. That is, it must satisfy:
\begin{align}
\label{eq:premature_inequality}
    \underbrace{\prod_{t'=1}^T p(y_{t'}|y_{<t'})}_{
    =p(\mathbf{y})}
    > 
    \underbrace{p(\left<\mathrm{eos}\right>|y_{< t}) \prod_{t'=1}^{t-1} p(y_{t'}|y_{<t'})}_{
    = p(\mathbf{y}_{\leq t})
    }
\end{align}
which is equivalent to
\begin{align*}
    \prod_{t'=t}^T p(y_{t'} | y_{<t'})
    >
    p(\left<\mathrm{eos}\right>|y_{<t}),
\end{align*}
because of the autoregressive formulation.

In order for this inequality to hold, the probability assigned to the $\eos$ must be extremely small, as the left-hand side of the inequality is the product of many probabilities. In other words, the dynamic range of the $\eos$ token probability must be significantly greater than that of any other token probability, in order for the autoregressive language model to properly capture the ill-formed nature of premature sequences. 

It is, however, a usual practice to treat the $\eos$ token just like any other token in the vocabulary, which is evident from Eq.~\eqref{eq:chainrule}. This leads to the difficulty in having a dramatically larger dynamic range for the $\eos$ probability than for other token probabilities. In other words, this limited dynamic range due to the lack of special treatment of $\eos$ is what previous studies have referred to as ``oversmoothing'', and this leads to the degeneracy in length modeling. 

Under this observation, we can now quantify the degree of oversmoothing\footnote{
To be strict, this should be called the degree of `smoothing', but we stick to oversmoothing to be in line with how this phenomenon has been referred to in previous studies.
} 
by examining how often the inequality in Eq.~\eqref{eq:premature_inequality} is violated:
\begin{definition}[Oversmoothing rate]
The oversmoothing rate of a sequence is defined as
\begin{align}
    \label{eq:ros}
    r_{\mathrm{os}}(\mathbf{y})
    =
    \frac{1}{|\mathbf{y}|-1}
    \sum_{t=1}^{|\mathbf{y}|-1}
    \mathds{1}
    &
    \Big(
    \prod_{t'=t}^{|\mathbf{y}|} p(y_{t'} | y_{<t'})
    \nonumber
    \\
    &<
    p(\eos|y_{<t})
    \Big),
\end{align}
where $\mathds{1}$ is an indicator function returning $1$ if true and otherwise $0$. 
\end{definition}
With this definition, we can now quantify the degree of oversmoothing and thereby quantify any improvement in terms of the issue of oversmoothing by any future proposal, including our own in this paper.

Because premature sequences may be well-formed, it is not desirable for the oversmoothing rate to reach $0$. We, however, demonstrate later empirically that this oversmoothing rate is too high for every system we considered in this work. 

\subsection{Minimizing the oversmoothing rate}
\label{ssec:osloss}

The oversmoothing rate above is defined as the average of indicator functions, making it challenging to directly minimize.
We instead propose to minimize an upper bound on the original oversmoothing rate, that is differentiable almost everywhere and admits gradient-based optimization:
\begin{definition}[Oversmoothing loss]
Given a sequence $\mathbf{y}$, the oversmoothing loss is defined as

\begin{align*}
    l_\mathrm{os}(\mathbf{y}) = \frac{1}{|\mathbf{y}|}\sum_{t=1}^{|\mathbf{y}|} &\max \Bigg(0, \log p(\eos|y_{<t})\\ &-\sum_{t^\prime=t}^{|\mathbf{y}|} \log p(y_{t^\prime} | y_{<t^\prime}) + m \Bigg),
\end{align*}
which is an upper bound of $r_{\mathrm{os}}(y)$ with $m \geq 1$.
\end{definition}

We use this oversmoothing loss as a regularization term and augment the original objective function with it. We use $\alpha \in [0, 1)$ to balance the relative strengths of these two terms:
\begin{align*}
    l(\mathbf{y}) = (1-\alpha) \cdot l_\mathrm{nll}(\mathbf{y}) + \alpha \cdot l_\mathrm{os}(\mathbf{y}),
\end{align*}
where $l_\mathrm{nll}(\mathbf{y}) = - \sum_{t=1}^{|\mathbf{y}|} \log p(y_t|y_{<t})$.

When the inequality in Eq.~\eqref{eq:premature_inequality} is satisfied at step $t$ with the log-probability difference between the l.h.s. and r.h.s. at least as large as $m$, the oversmoothing loss disappears, implying that the step $t$ does not contribute to the issue of oversmoothing. When this loss is activated at step $t$, we have two terms, excluding the constant margin $m$, the log-probability of {\it incorrect} $\eos$ given the context $y_{<t}$ and the negative log-probability of the {\it correct} suffix given the same context.

Minimizing the first term explicitly prevents a premature sequence $\mathbf{y}_{\leq t}$ from being a valid sequence by lowering the probability $y_t$ being $\left<\mathrm{eos}\right>$ even further compared to the other tokens in the vocabulary. The second term on the other hand prevents the premature sequence by ensuring that the full sequence $\mathbf{y}=(y_{<|y|}, \left<\mathrm{eos}\right>)$ is more likely than the premature sequence $\mathbf{y}_{\leq t} = (y_{<t}, \left<\mathrm{eos}\right>)$. In short, the proposed oversmoothing loss addresses both of these scenarios which lead to oversmoothing. Finally, only when both of these factors are suppressed enough, the loss vanishes.

The second scenario above, i.e., increasing the probability of a suffix at each position $t$, has the effect of greatly emphasizing the latter part of the sequence during training. This can lead to a degenerate case in which the earlier part of a sequence cannot be modeled by an autoregressive sequence modeling, if the strength of the proposed oversmoothing loss is too large. We thus use this loss together with the original negative log-likelihood loss ($\alpha > 0$) only after pretraining a model with the negative log-likelihood loss only ($\alpha=0$).

\section{Related work}
\label{sec:related_work}

The issue of generating sequences that are shorter than the ground-truth one has been studied from various aspects including model parametrization, data collection, and decoding. Here we highlight some of these projects in the context of our work.

On the aspect of model parametrization, 
\citet{peters2021smoothing} suggest using sparse transformation of the next-token distribution rather than the usual way of using softmax. Such a model is then able to assign zero probability to short sequences more readily and thereby reduce the oversmoothing rate. Their approach, however, does not explicitly encourage $\eos$ tokens to be assigned zero probability, unlike ours where $\eos$ is treated specially.
\citet{shi2020neural} embed the $\eos$ token with a distinct vector at each position within the sequence. This was shown to help the probability of empty sequence, although they do not report its impact on translation quality at all. 

On data collection, \citet{nguyen2021data} analyze data collection and show that data augmentation techniques altering sequence length may address the issue of oversmoothing and improve translation quality. Their work is however limited to low-resource tasks. With respect to decoding, \citet{murray2018correcting} design a decoding algorithm that learns to correct the underestimated length. Alternative decoding algorithms, such as minimum Bayes risk decoding \citep{eikema2020map, mller2021understanding}, have been shown to alleviate the length mismatch to a certain extent when compared to beam search. 

These earlier approaches do not attempt at formally characterizing the cause behind the issue of oversmoothing. This is unlike our work, where  we start by formalizing the issue of oversmoothing and propose a way to alleviate this issue by directly addressing this cause.

\section{Experimental Setup}
\label{sec:exp_settings}

We follow a standard practice to train our neural machine translation models, following \citep{Ott_2018}, using FairSeq framework \citep{ott2019fairseq}. We use BPE tokenization via either fastBPE \cite{sennrich2016bpe} or SentencePiece \cite{kudo2018sentencepiece}, depending on the dataset.
Although it is not required for us to use state-of-the-art models to study the issue of oversmoothing, 
we use models that achieve reasonable translation quality.
The code implementing FairSeq task with the oversmoothing rate metric, oversmoothing loss, and experimental results is available on Github.\footnote{
\url{https://github.com/uralik/oversmoothing_rate}
}

\subsection{Tasks and Models}

We experiment with both smaller datasets using language pairs from IWSLT'17 and larger datasets using language pairs from WMT'19 and WMT'16. In the latter case, we use publicly available pretrained checkpoints in FairSeq. We execute five training runs with different random initialization for every system. These language pairs and checkpoints cover different combinations of languages and model sizes.
This allows us to study the oversmoothing rate under a variety of different settings. 

\paragraph{IWSLT'17 \{De,Fr,Zh\}$\rightarrow$En:}

We adapt the data preprocessing procedure from FairSeq IWSLT recipe and use SentencePiece tokenization. The training sets consist of 209K, 236K, and 235K sentence pairs for De$\rightarrow$En, Fr$\rightarrow$En, and Zh$\rightarrow$En, respectively. We use the TED talks 2010 development set for validation, and the TED talks 2010-2015 test set for testing. The development and test sets, respectively, consist of approximately 800 and 8,000 sentence pairs for all tasks. 

We use the same architecture named \texttt{transformer\_iwslt\_de\_en} in FairSeq for each language pair. It consists of 6 encoder and decoder layers with 4 self-attention heads followed by feed-forward transformations. Both encoder and decoder use embeddings of size 512 while the input and output embeddings are not shared. Both the encoder and decoder use learned positional embedding. We early-stopping training based on the validation set. Evaluation is done on the test set.

\paragraph{WMT'16 En$\rightarrow$De:}

We prepare the data following the recipe from FairSeq Github.\footnote{
\url{https://git.io/JDqB2}
}
The training set has 4.5M sentence pairs. 
Following \citet{ott2018scaling}, we use newstest13 as the development set and newstest14 as the test set, they contain 3K sentence pairs each. 
We fine-tune the pretrained checkpoint which was originally released by \citep{ott2018scaling} and is available from FairSeq. The recipe uses a transformer architecture based on \cite{vaswani17attention}. Different from all other models considered in this work, this architecture shares vocabulary embeddings between the encoder and the decoder.

\paragraph{WMT'19 Ru$\rightarrow$En, De$\leftrightarrow$En}

We closely follow \citet{Ng_2019} in preparing data, except for filtering based on language identification. We use the subset of WMT'19 training set consisting of news commentary v12 and common crawl resulting in slightly more than 1M and 2M training sentence pairs for Ru$\rightarrow$En and De$\leftrightarrow$En pairs, respectively. 
We fine-tuned single model checkpoints from \citet{Ng_2019}.\footnote{
\url{https://git.io/JDqBo}
} 
We early-stop training on the official WMT'19 development set. For evaluation, we use the official WMT'19 test set. 

\subsection{Training}

We use Adam optimizer \cite{Kingma15adam} with $\beta_1 = 0.9$ and $\beta_2=0.98$. We use the inverse square root learning scheduler with 4,000 warm-up steps. We use the initial learning rate of $5 \times 10^{-4}$, dropout rate of 0.3~\cite{srivastava14dropout} , and weight decay with its rate set to $10^{-4}$.
We use label smoothing with $0.1$ of probability smoothed uniformly during pretraining with NLL loss and turn it off after starting to use the oversmoothing loss.
We vary the oversmoothing loss weight $\alpha$ from $0.0$ to $0.95$ with a step size of $0.05$. We use a fixed margin $m=10^{-4}$ whenever we use the oversmoothing loss.

\paragraph{Early stopping}

We use early stopping for model selection based on the value of the objective function computed on the development set. We evaluate the model on the development set every 2K updates for IWSLT ($\sim$2K tokens per update) and WMT ($\sim$9K tokens per update) systems. We stop training when the objective has not improved over more 5 consecutive validation runs. 
We fine-tune models around 5K updates for IWSLT'17 DE-EN and ZH-EN, and 7K updates for IWSLT'17 FR-EN. As for WMT'19, it takes approximately 45K updates for DE-EN and EN-DE language pairs to early-stop, and 76K updates for RU-EN model, and 12K updates for WMT'16.

\subsection{Decoding}

To test translation quality, we translate a test set 
with beam search decoding, as 
implemented in the FairSeq. We vary beam sizes to study their effect in-depth. 
We set the lower- and upper-bound of a generated translation to be, respectively, 0 and $1.2\cdot l_x + 10$, where $l_x$ is the length of the source $x$.
We do not use either length normalization nor length penalty, in order to study the impact of oversmoothing on decoding faithfully.
We compute and report BLEU scores using \texttt{sacreBLEU} on detokenized predictions.

\begin{figure}[t]
    \centering
    \includegraphics[width=1.\linewidth]{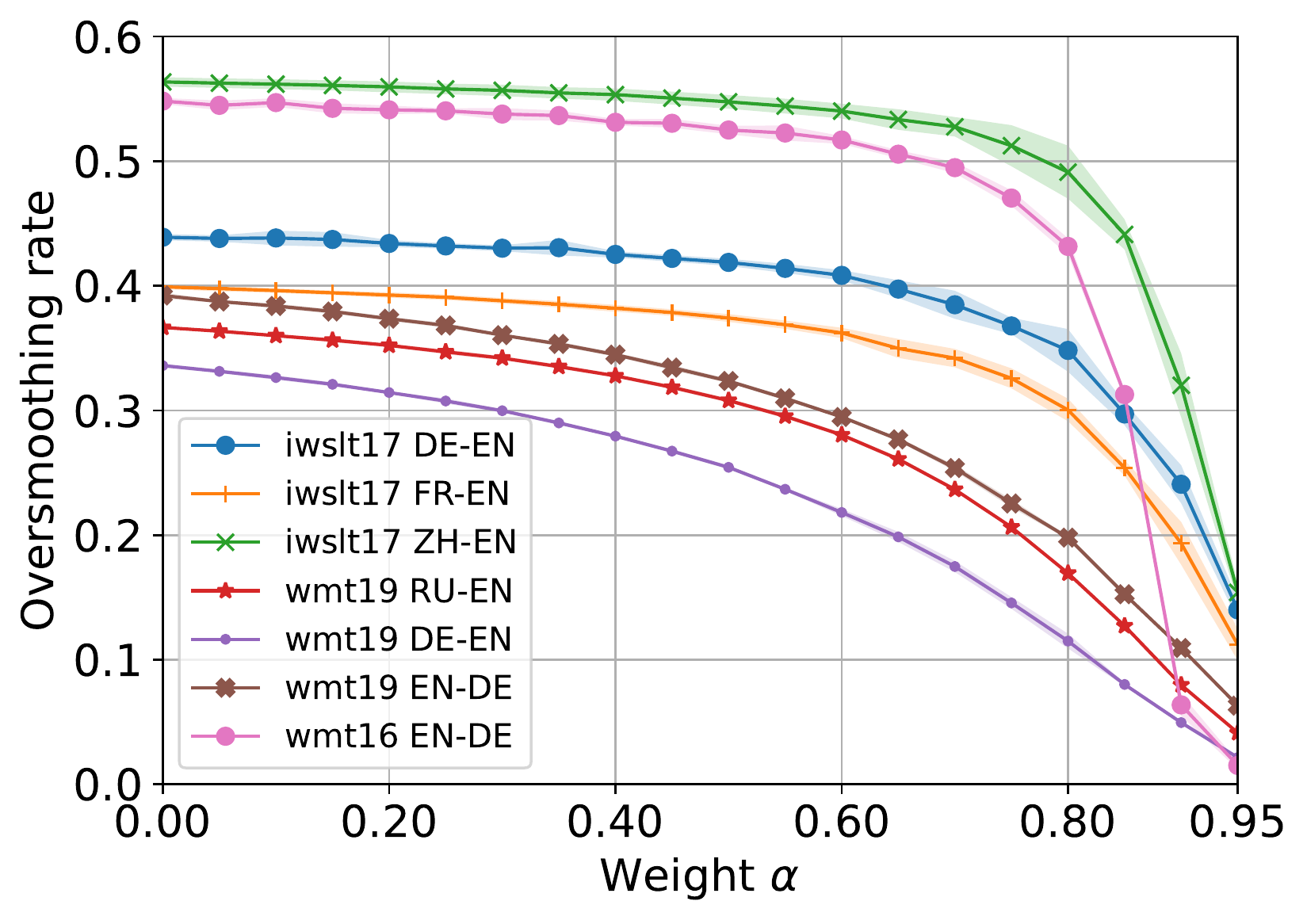}
    \caption{Average oversmoothing rate is going down as we increase contribution of the oversmoothing loss during fine-tuning. Filled regions denote the standard deviation across training runs according to \cref{sec:exp_settings}.}
    \label{fig:osrate_vs_alpha}
\end{figure}

\begin{figure*}[ht]
    \centering
    \begin{minipage}{0.48\linewidth}
    \centering
    \includegraphics[width=1.\linewidth]{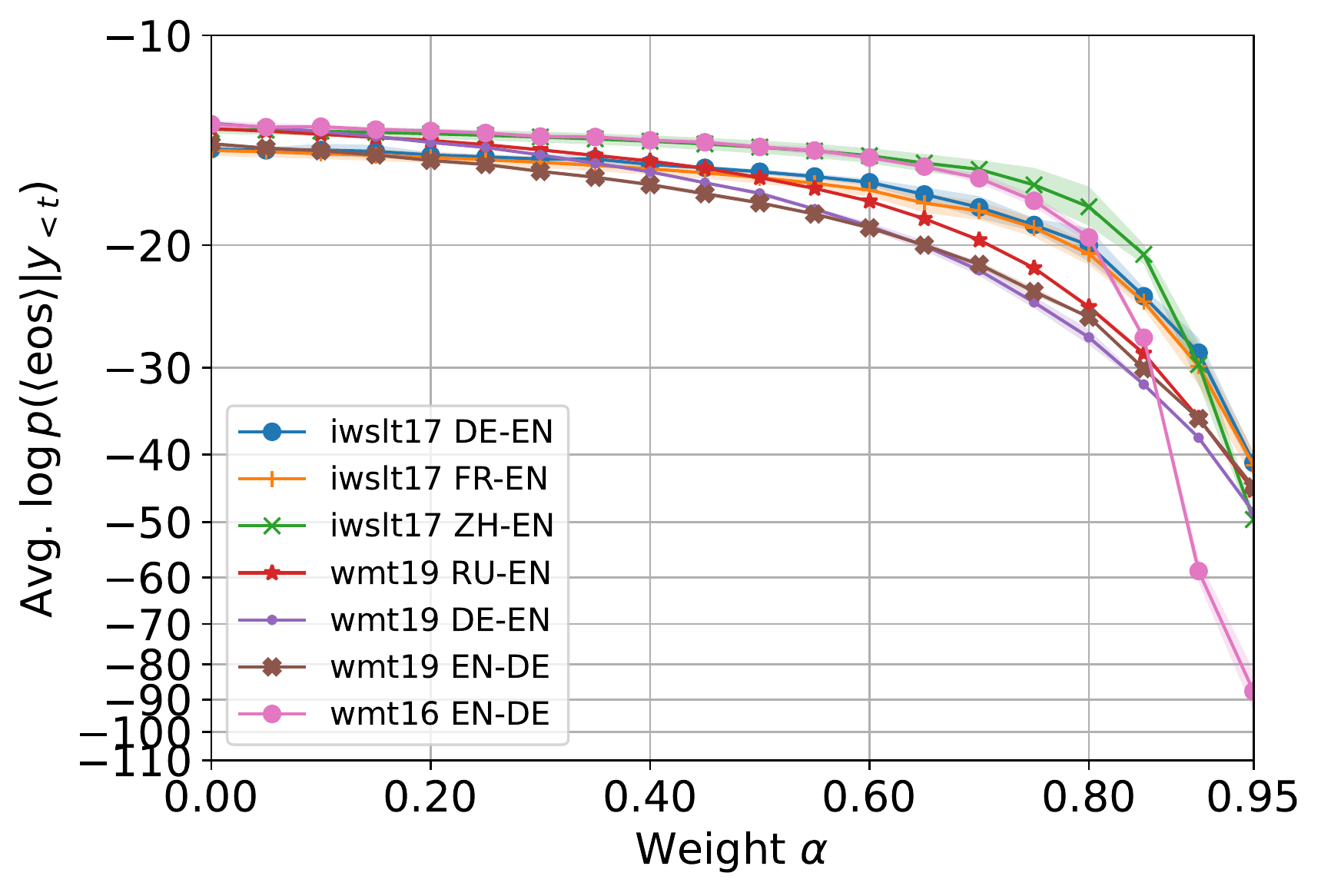}
    
    (a)
    \end{minipage}
    \hfill
    \begin{minipage}{0.48\linewidth}
    \centering
    \includegraphics[width=0.97\linewidth]{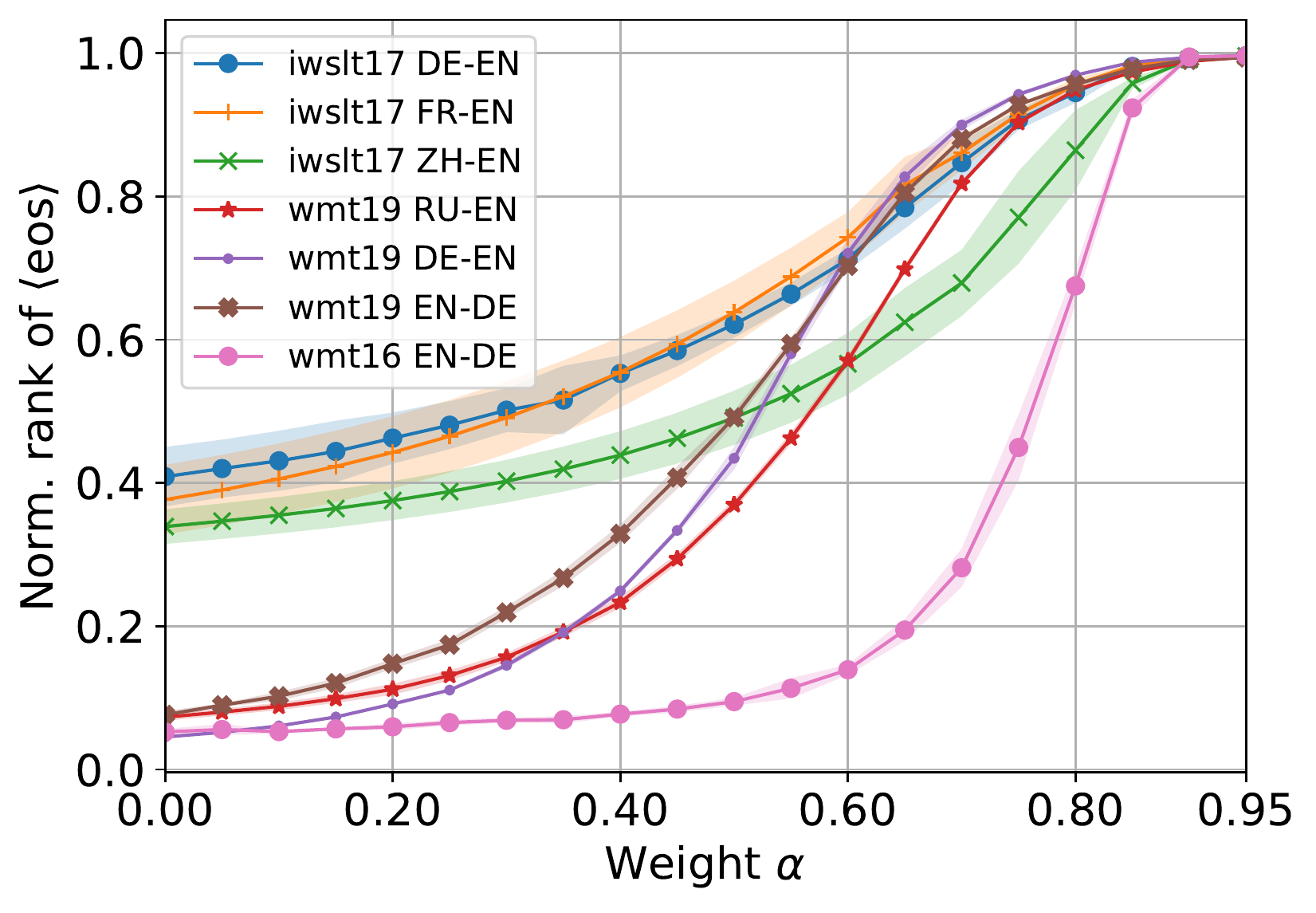}
    
    (b)
    \end{minipage}
    
    \caption{(a) Log-probabilities of $\eos$ token within length-$t$ prefixes averaged across all positions per translation and then averaged across all translations. 
    (b) Normalized rank of $\eos$ token within length-$t$ prefixes averaged across all positions $t$ per translation and then averaged across all translations. $1$ means the lowest rank within the vocabulary. Filled regions denote the standard deviation across training runs according to \cref{sec:exp_settings}.}
    \label{fig:logp_eos}
\end{figure*}

\section{Experiments}
\label{sec:experiments}

As we pointed out earlier, publicly available translation systems 
exhibit a high degree of oversmoothing.
See the left-most part of \cref{fig:osrate_vs_alpha}, where $\alpha=0$. In particular, this rate ranges from $\textbf{34\%}$ (WMT'19 DE$\rightarrow$EN) up to $\textbf{56\%}$ (IWSLT'17 ZH$\rightarrow$EN).

According to \cref{ssec:osloss}, the oversmoothing rate should decrease as we increase the relative strength of the oversmoothing loss. To verify this, we fine-tune these models while varying the coefficient $\alpha$. In \Cref{fig:osrate_vs_alpha} we demonstrate the oversmoothing rate reduces all the way down to $\textbf{3\%}$ (WMT'19 DE$\rightarrow$EN)
and $\textbf{17\%}$ (IWSLT'17 ZH$\rightarrow$EN)
as we increase the strength of the regularizer. 
The oversmoothing rate monotonically decreases for every system we consider, as we increase $\alpha$ up to $0.95$.

\subsection{Regularization and $\eos$ token}

Minimizing the proposed oversmoothing loss minimizes the log-probability of $\eos$ token at the end of every length-$t$ prefix unless it is already low enough. We analyze how the strength of regularization affects the average log-probability of $\eos$ token measured at the end of each premature translation. As presented in \Cref{fig:logp_eos}~(a), the log-probability of $\eos$ at the end of premature sequences decreases monotonically as the oversmoothing rate decreases (i.e., as the strength of the oversmoothing loss increase, as evident from \Cref{fig:osrate_vs_alpha}).

Although the log-probability of $\eos$ is an important factor in oversmoothing, \citet{welleck2020consistency} claim that it is the rank of $\eos$ token that matters when using an incomplete approximate decoding strategy, such as beam search, for generation. We thus look at the average normalized rank of $\eos$ token at the end of every length-$t$ prefix in \Cref{fig:logp_eos}~(b). The rank drops rapidly and almost monotonically as we add more regularization. The effect of regularization is more visible with the rank than with the log-probability, especially when $\alpha$ is small.

\begin{figure}[t]
    \centering
    \includegraphics[width=1.\linewidth]{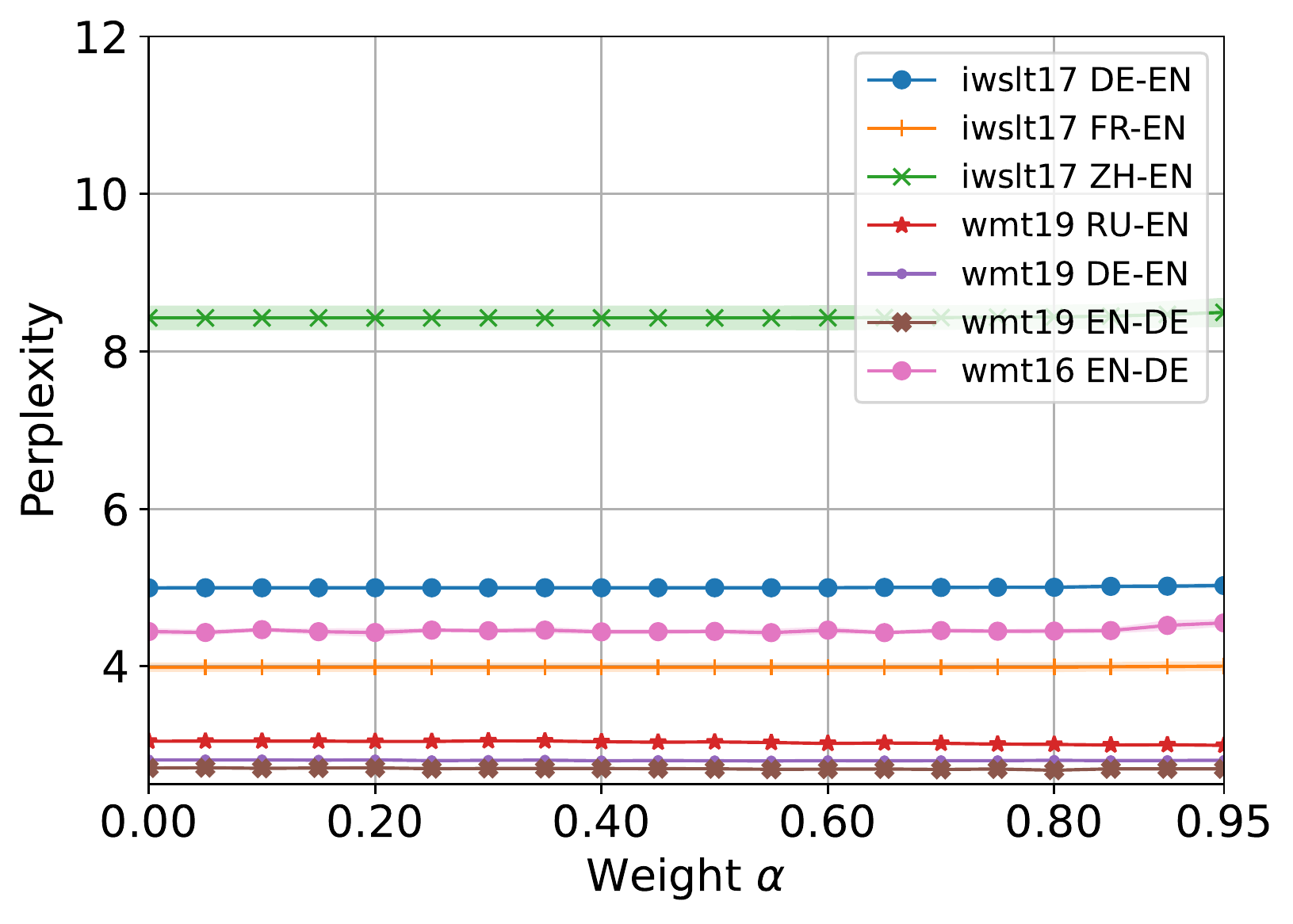}
    \caption{Perplexity measured on reference translations remains stable as we increase the strength of the regularization. Filled regions denote the standard deviation across training runs according to \cref{sec:exp_settings}.}
    \label{fig:ppls}
\end{figure}

Although the proposed regularization reduces the probability of $\eos$ token where it is not supposed to be, we observe that the performance of the system as a language model does not degrade much regardless of the chosen value of $\alpha$. This is evident from the flat lines in \Cref{fig:ppls} where we plot the perplexity of each model while varying $\alpha$. This demonstrates that there are many different ways to minimize the negative log-likelihood, and some of those solutions exhibit a higher level of oversmoothing than the others. The proposed oversmoothing loss is an effective way to bias the solution toward a lower level of oversmoothing.

\begin{figure*}[!ht]
     \centering
     \begin{subfigure}[b]{0.49\textwidth}
         \centering
         \includegraphics[width=\textwidth]{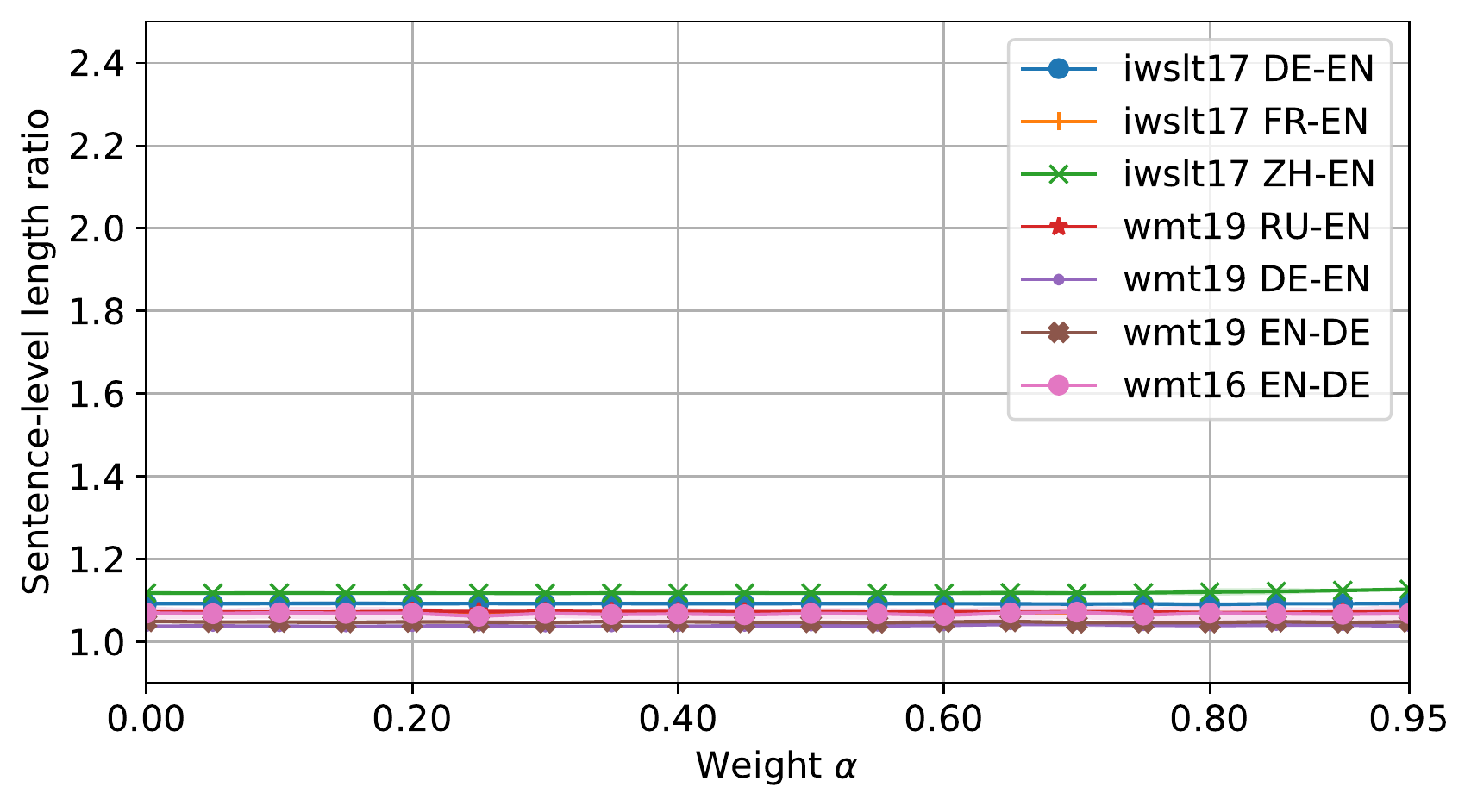}
         \caption{beam 5}
     \end{subfigure}
    \begin{subfigure}[b]{0.49\textwidth}
         \centering
         \includegraphics[width=\textwidth]{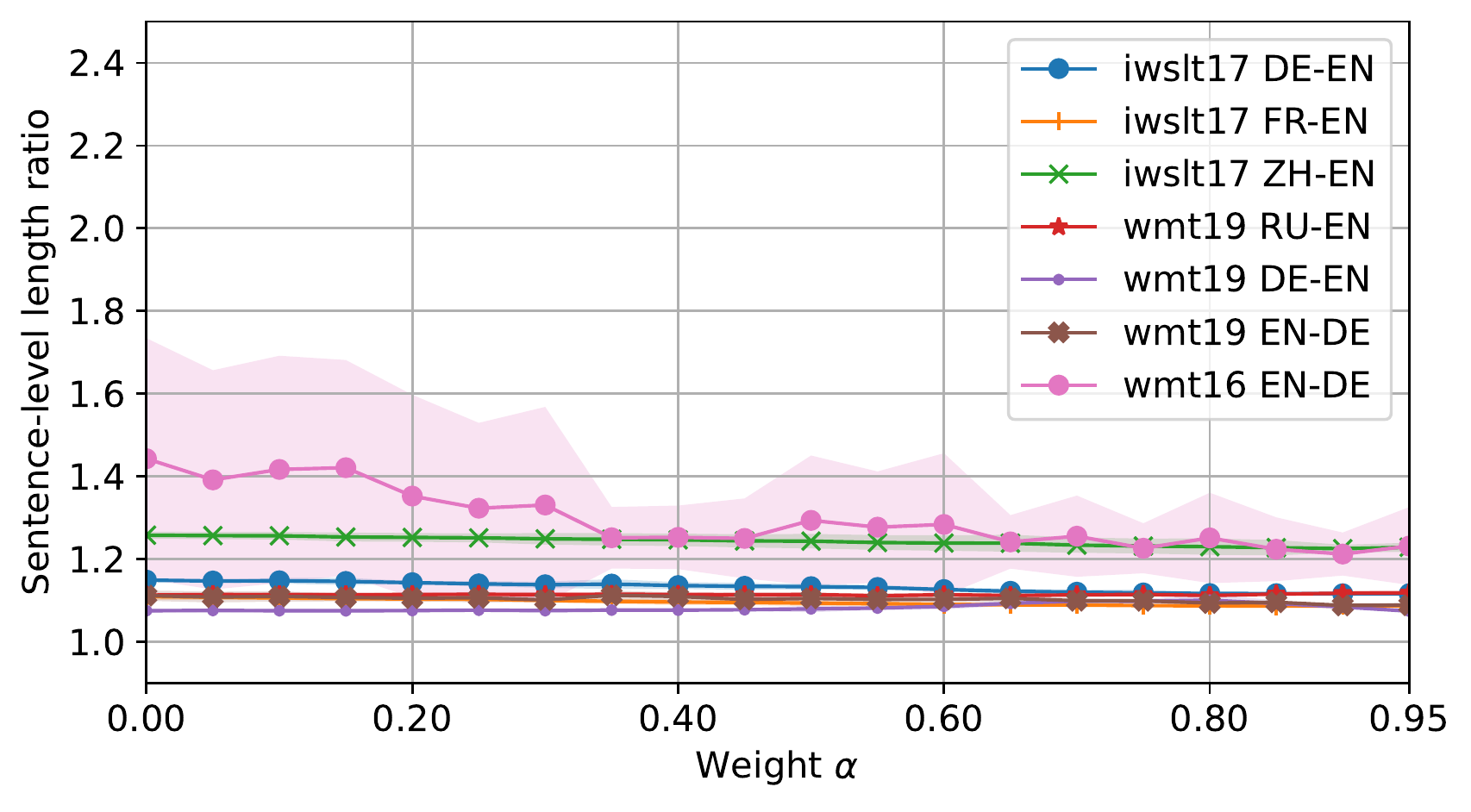}
         \caption{beam 250}
     \end{subfigure} \\
    \begin{subfigure}[b]{0.49\textwidth}
         \centering
         \includegraphics[width=\textwidth]{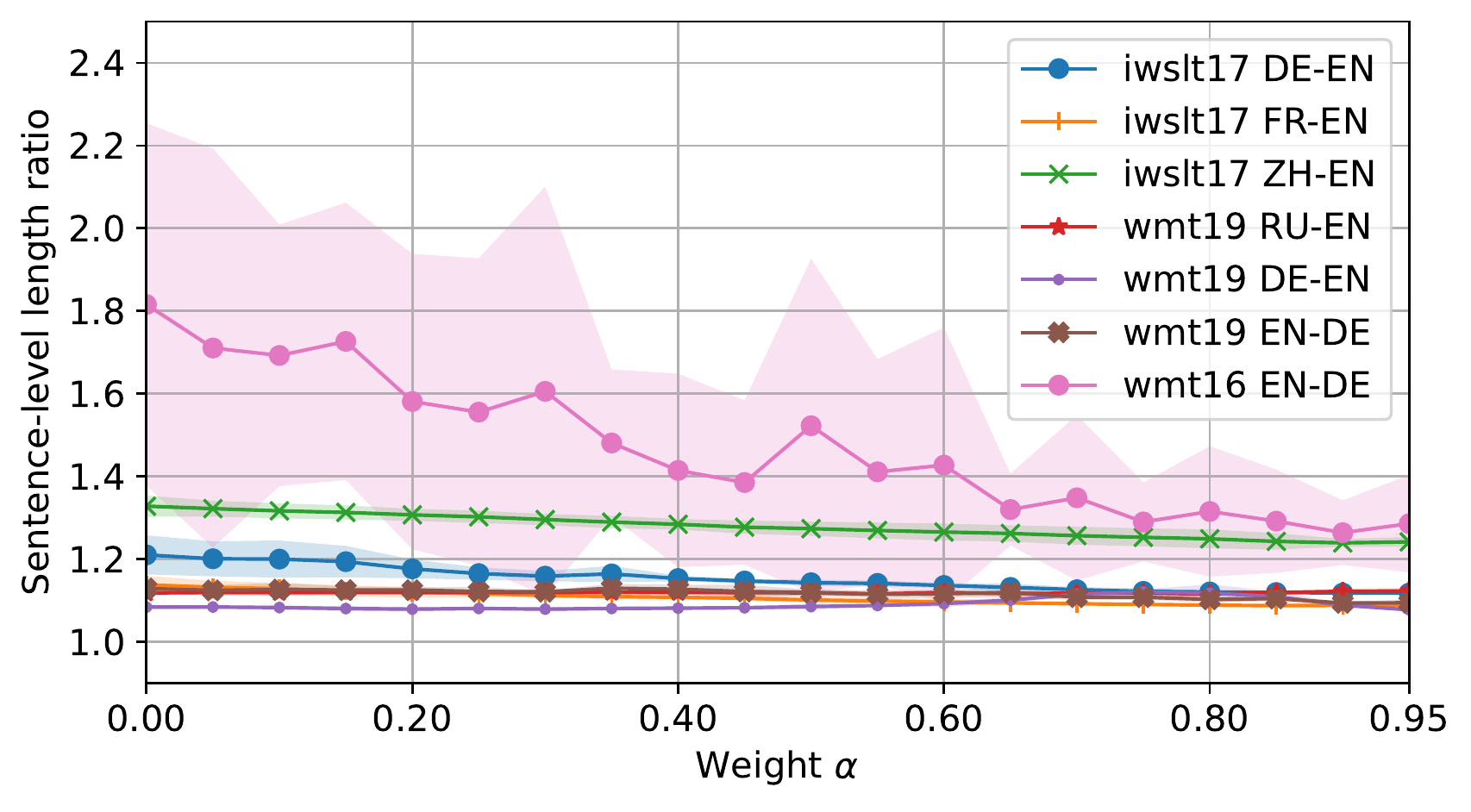}
         \caption{beam 500}
     \end{subfigure}
    \begin{subfigure}[b]{0.49\textwidth}
         \centering
         \includegraphics[width=\textwidth]{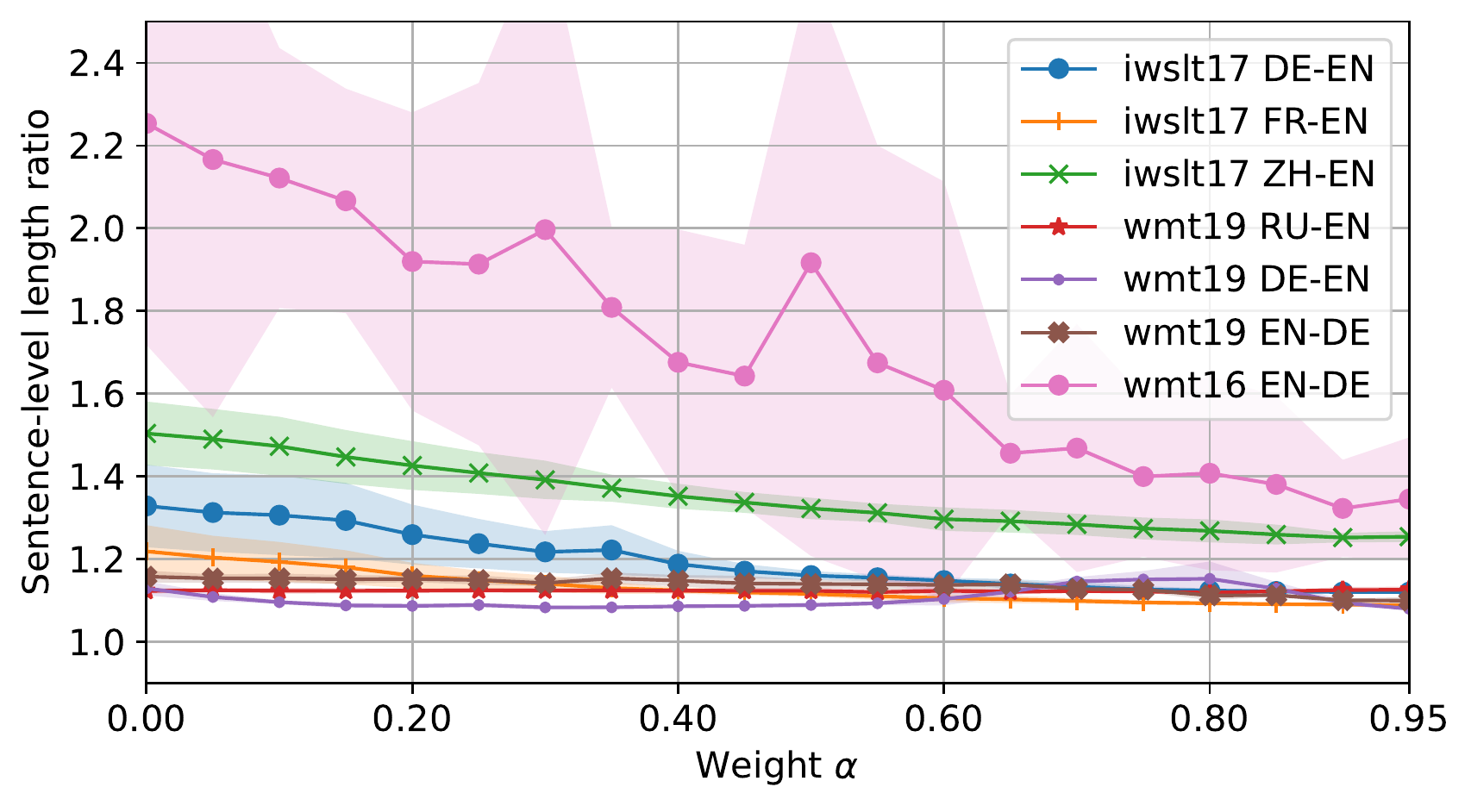}
         \caption{beam 1000}
     \end{subfigure}
    \caption{Sentence-level length ratio is $ \frac{1}{|D_\mathrm{test}|} \sum_{i=1}^{|D_\mathrm{test}|}  |\mathbf{y}^{\mathrm{ref}}_{i}| / |\mathbf{y}^{\mathrm{beam}}_{i}|$, where $\mathbf{y}^{\mathrm{beam}}_{i}$ is generated translation using beam search for $i$-th input sentence from the test set $D_\mathrm{test}$, and $\mathbf{y}^{\mathrm{ref}}_{i}$ is the corresponding reference translation. Filled regions denote the standard deviation across training runs according to \cref{sec:exp_settings}.}
    \label{fig:sent_length_ratio}
\end{figure*}

\subsection{Oversmoothing rate and decoding}

\begin{figure*}[!ht]
     \centering
     \begin{subfigure}[b]{0.32\textwidth}
         \centering
         \includegraphics[width=\textwidth]{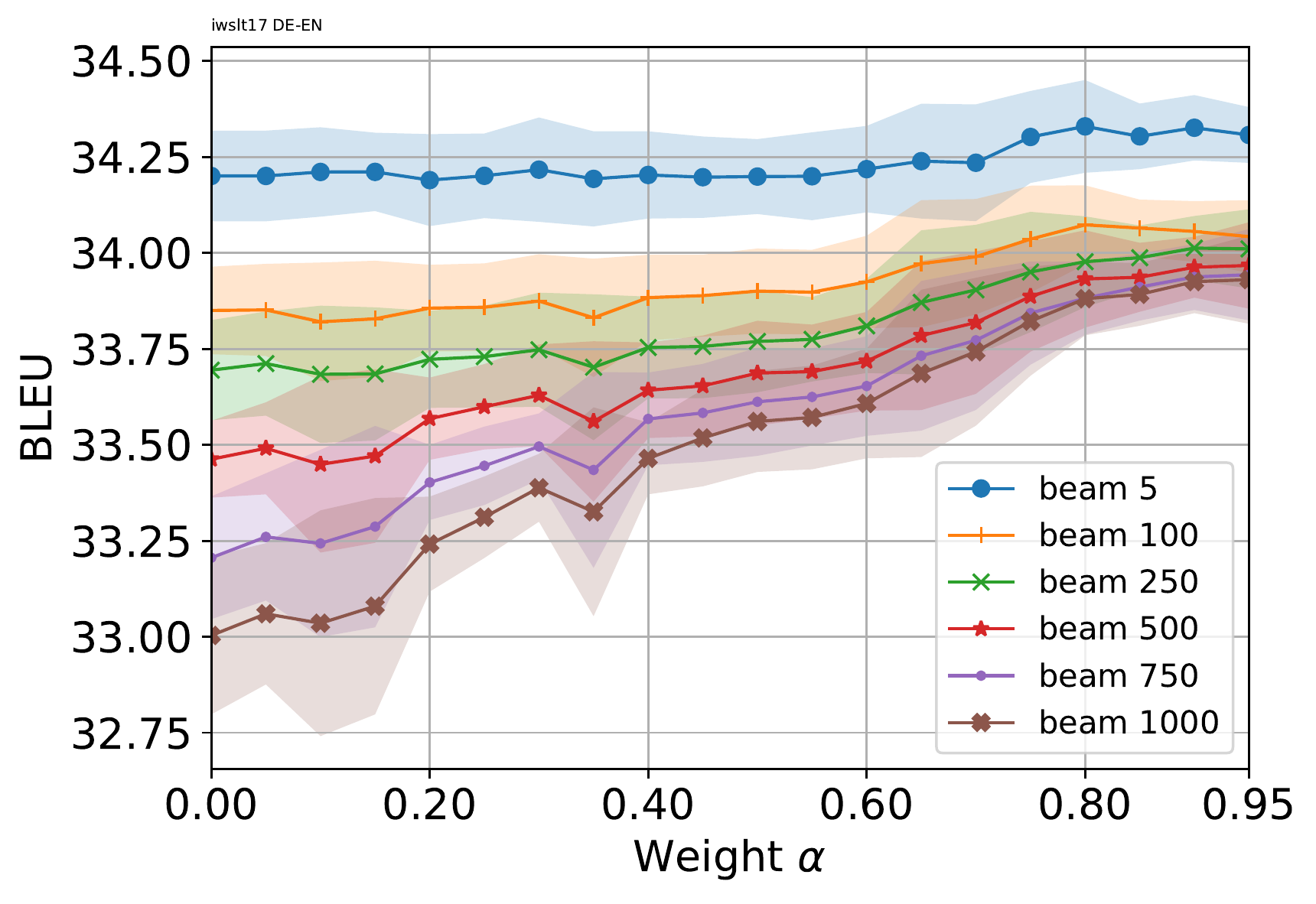}
         \caption{IWSLT'17 DE$\rightarrow$EN}
     \end{subfigure}
    \begin{subfigure}[b]{0.31\textwidth}
         \centering
         \includegraphics[width=\textwidth]{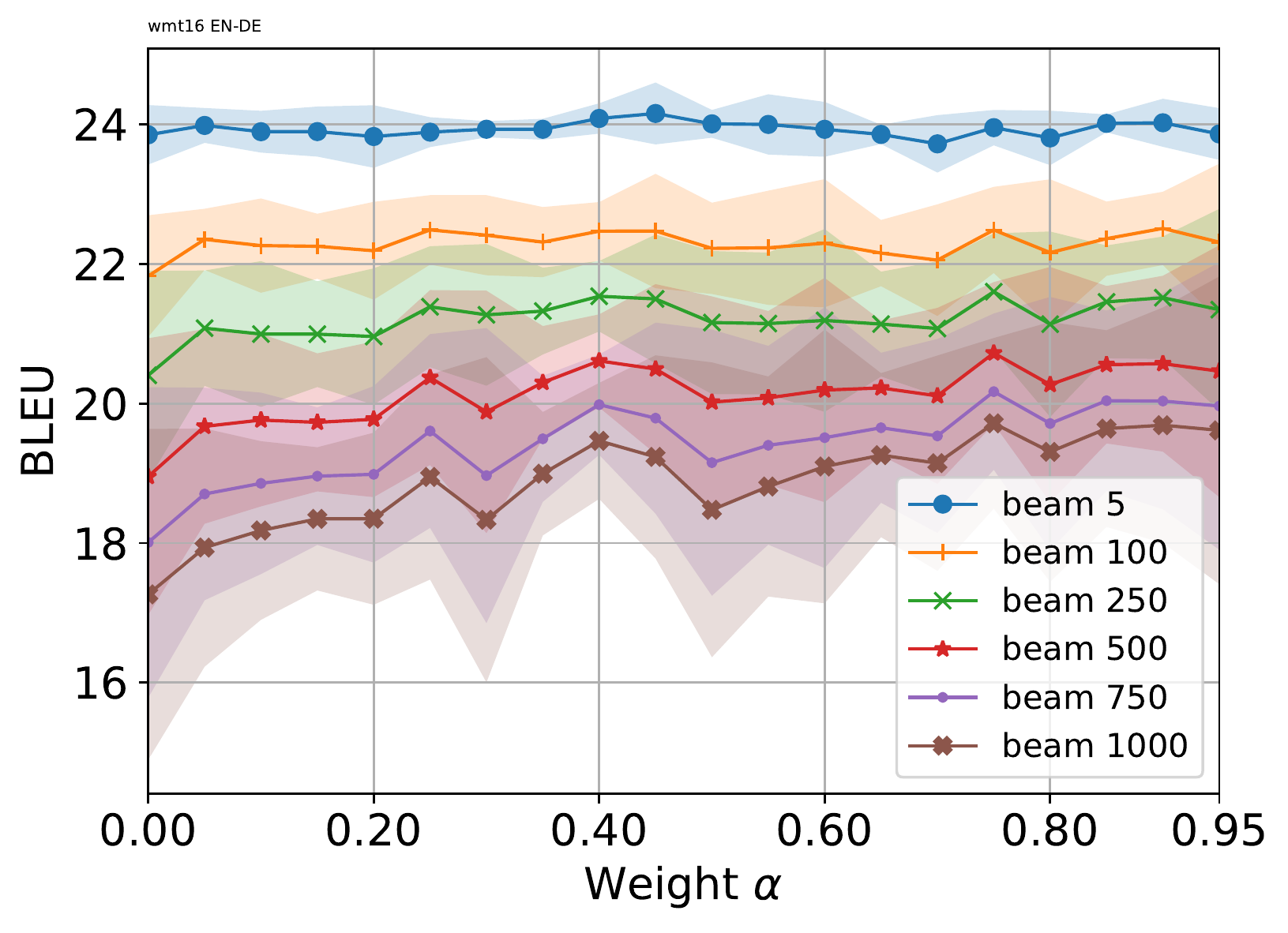}
         \caption{WMT'16 EN$\rightarrow$DE}
     \end{subfigure}
    \begin{subfigure}[b]{0.31\textwidth}
         \centering
         \includegraphics[width=\textwidth]{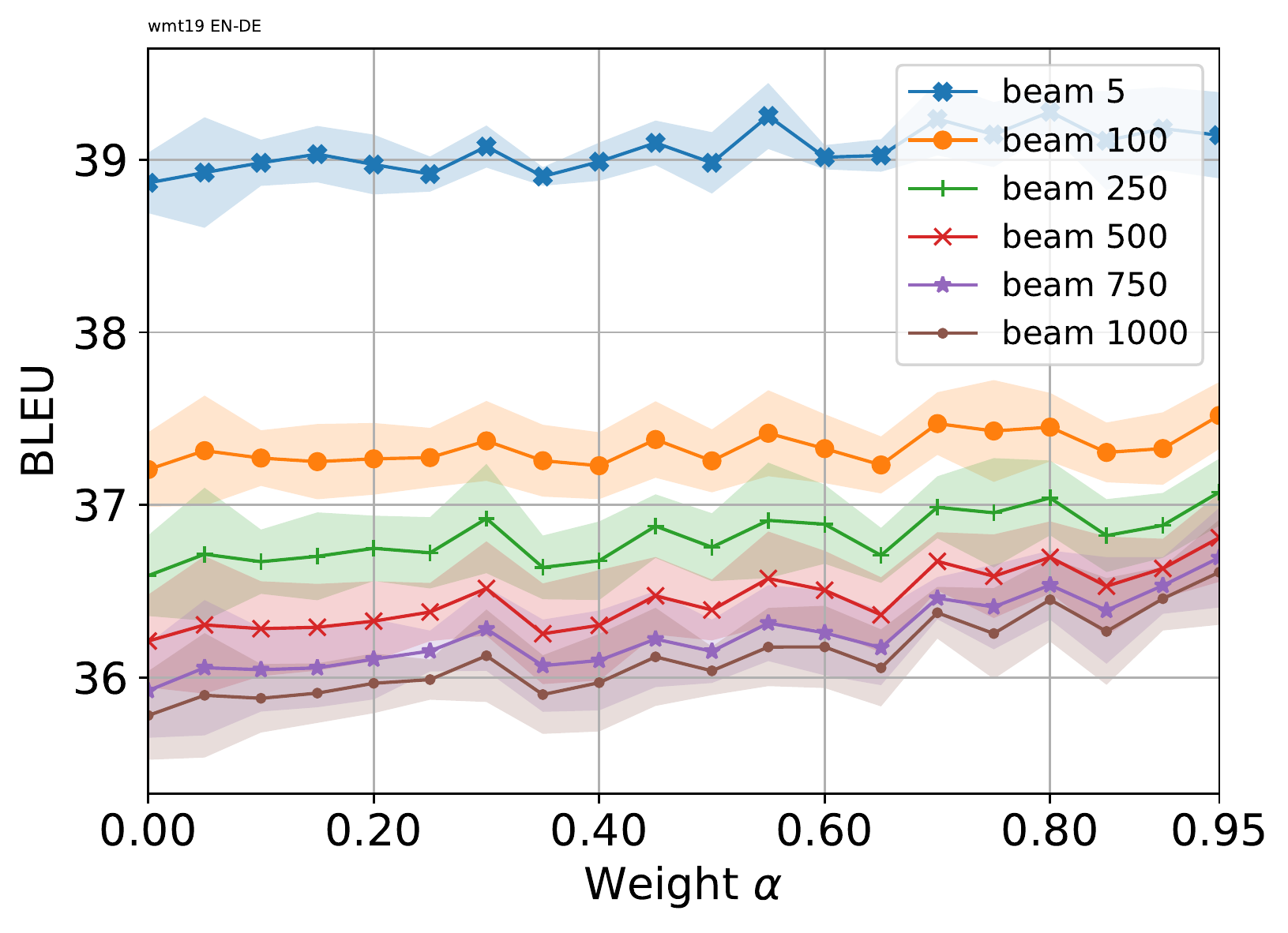}
         \caption{WMT'19 EN$\rightarrow$DE}
     \end{subfigure}
    \caption{BLEU score is measured on corresponding test sets. Decoding is done using beam search with beam sizes given in the legend. \cref{sec:exp_settings} provides more details on test sets and decoding hyper-parameters. Filled regions denote the standard deviation across training runs according to \cref{sec:exp_settings}.}
    \label{fig:bleu_three_systems}
\end{figure*}

Earlier \citet{koehn2017six} noticed this issue of oversmoothing by observing that the length of generated sequences dramatically dropped as the beam width increased.
We confirm the decreasing length of generated translation as the beam size increases in \cref{fig:sent_length_ratio} when $\alpha = 0$. We study the change of this length as we add more regularization and calculate the sequence-level length ratio in \cref{fig:sent_length_ratio}. We observed a similar trend with other beam sizes (see \cref{sec:extra_results} for more details).

When fine-tuned with the proposed oversmoothing loss, the length ratio degrades significantly less, as we increase the beam size during decoding, than without. For instance, with $\alpha \geq 0.8$ the length ratio remains more or less constant with respect to the size of the beam. Despite the observed robustness, decoding with a smaller beam size produces translations with lengths which match reference lengths better regardless of the strength of regularization. 

\paragraph{Translation quality}

The quality of the produced translation is directly related to its length, because this length needs to closely match the length of the reference translation. However, the length information is not sufficient to make a conclusion about the translation quality. We quantify the quality of the translation 
by calculating the corpus-level BLEU score. 
The scores in \cref{fig:bleu_three_systems} indicate that the reduced degradation of length modeling does correlate with the improvements in translation quality, although the degree of such correlation varies across language pairs and beam widths. We highlight two major aspects of the effect of regularization on the translation quality. First, the impact of regularization is only visible when the beam size is substantially larger than what is commonly used in practice. Second, the degradation of translation quality with a larger beam size lessens as oversmoothing does as well, but it does not eliminate the degradation fully. Similar observations hold for all the other cases as well, which we present in \cref{sec:extra_results}. These observations imply that the effectiveness of approximate decoding in neural machine translation remains unsolved, despite our successful attempt at addressing the issue of oversmoothing. 

\section{Conclusion}

In this work, we tackled a well-reported issue of oversmoothing in neural autoregressive sequence modeling, which has evaded rigorous characterization until now despite of its ubiquity.
We characterized it by defining the oversmoothing rate. It computes how often the probability of the ground-truth sequence is lower than the probability of any of its prefixes.
We confirmed that the oversmoothing rate is too high among well-trained neural machine translation systems and proposed a way to directly minimize it during training.
We designed a differentiable upper bound of the oversmoothing rate called the oversmoothing loss. We experimented with a diverse set of neural machine translation systems to study the effect of the proposed regularization.

The experiments revealed several findings and takeaways. First, the oversmoothing loss is effective: we were able to monotonically decrease the oversmoothing rate by increasing the strength of the loss. 
Second, we found that this regularization scheme significantly expands the dynamic range of the log-probability of $\eos$ token and has even greater impact on its rank, without compromising on sequence modeling.
Third, the proposed approach dramatically alters the behaviour of decoding when a large beam width was used. More specifically, it prevents the issue of degrading length ratio and improves translation quality. These effects were not as apparent with a small beam size though.
The proposed notion of oversmoothing explains some of the degeneracies reported earlier, and the proposed mitigation protocol alleviates these degeneracies. We, however, find that the proposed approach could not explain a more interesting riddle, that is, the lack of improvement in translation quality despite lower oversmoothing when beam search with a smaller beam was used. This unreasonable effectiveness of beam search continues to remain a mystery and needs to be investigated further in the future.

\section*{Acknowledgements}
This work was supported by Samsung Advanced Institute of Technology (under the project \textit{Next Generation Deep Learning: From Pattern Recognition to AI}) and NSF Award 1922658 NRT-HDR: FUTURE Foundations, Translation, and Responsibility for Data Science.

\bibliography{anthology,custom}
\bibliographystyle{acl_natbib}

\clearpage

\appendix

\section{Additional results for translation quality experiment}
\label{sec:extra_results}

\begin{figure}[!ht]
     \centering
     \begin{subfigure}[b]{1.\linewidth}
         \centering
         \includegraphics[width=\linewidth]{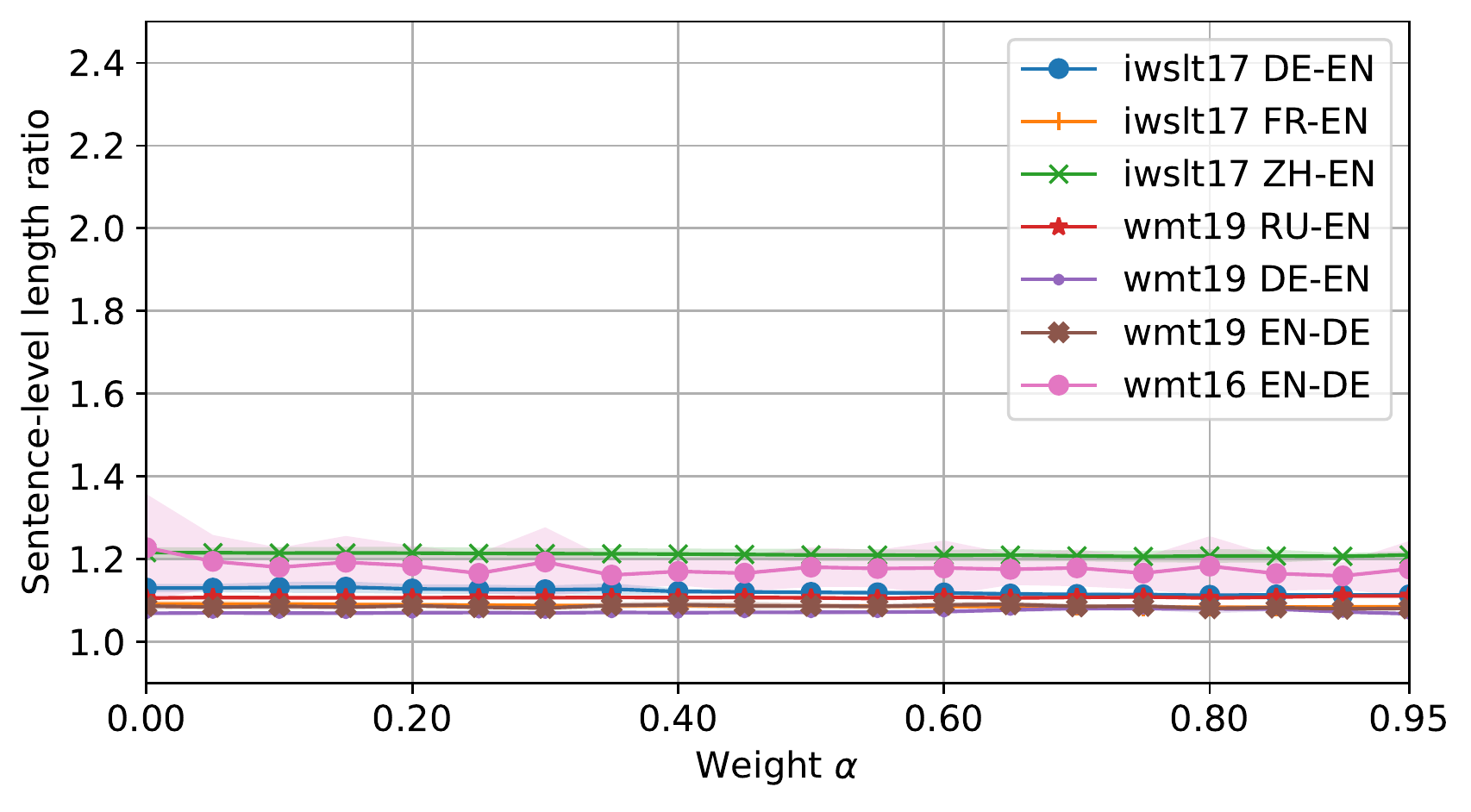}
         \caption{beam 100}
     \end{subfigure}
    \begin{subfigure}[b]{1,\linewidth}
         \centering
         \includegraphics[width=\linewidth]{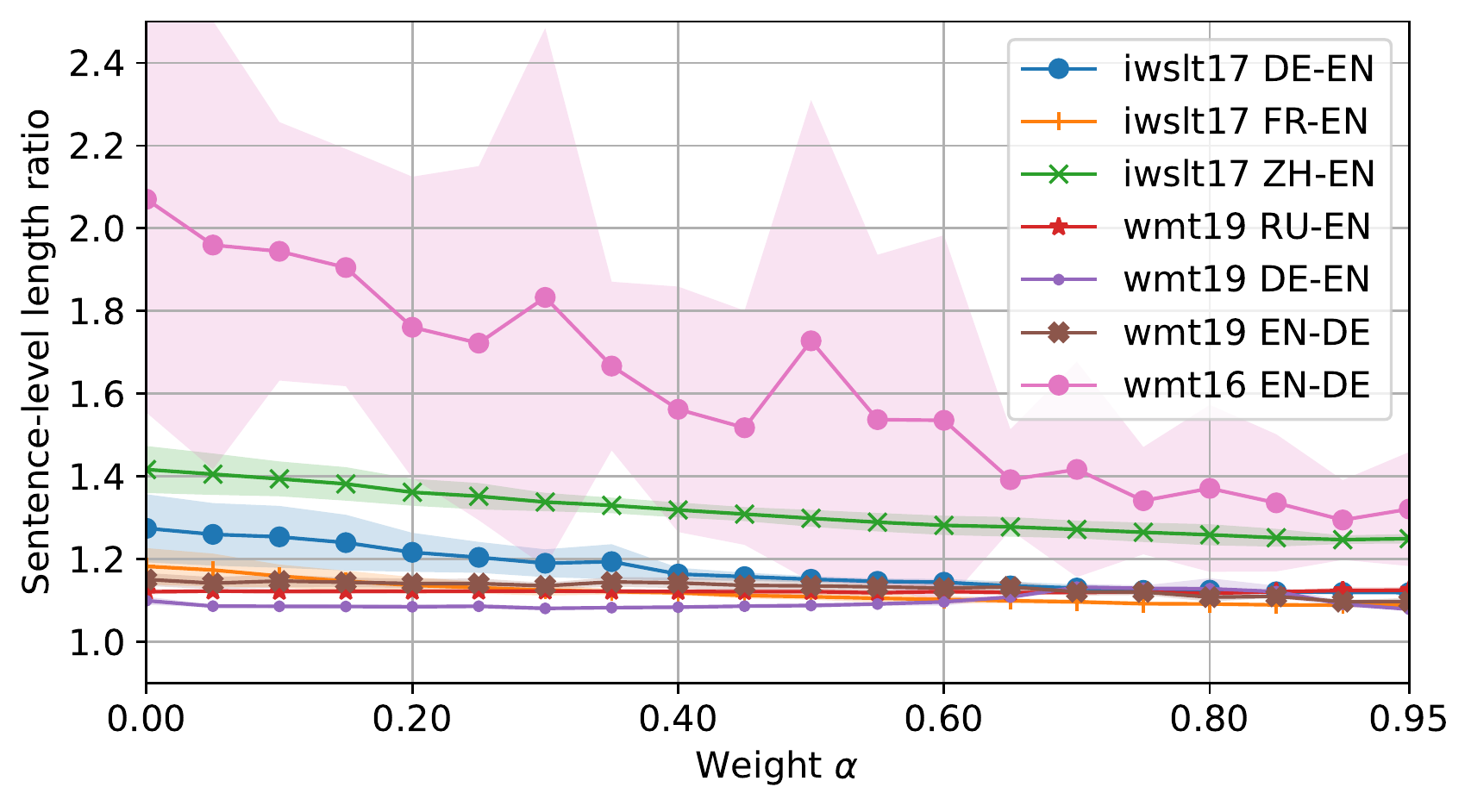}
         \caption{beam 750}
     \end{subfigure}
    \caption{Sentence-level length ratio is $ \frac{1}{|D_\mathrm{test}|} \sum_{i=1}^{|D_\mathrm{test}|} |\mathbf{y}^{\mathrm{ref}}_{i}| / |\mathbf{y}^{\mathrm{beam}}_{i}|$, where $\mathbf{y}^{\mathrm{beam}}_{i}$ is generated translation using beam search for $i$-th input sentence from the test set $D_\mathrm{test}$, and $\mathbf{y}^{\mathrm{ref}}_{i}$ is the corresponding reference translation. Filled regions denote the standard deviation across training runs according to \cref{sec:exp_settings}.}
    \label{fig:sent_length_ratio_appendix}
\end{figure}

\cref{fig:sent_length_ratio_appendix} presents sentence-level length ratio values for decoding settings which are not reported in the main text.

\begin{figure*}[!ht]
     \centering
     \begin{subfigure}[b]{.49\linewidth}
         \centering
         \includegraphics[width=\linewidth]{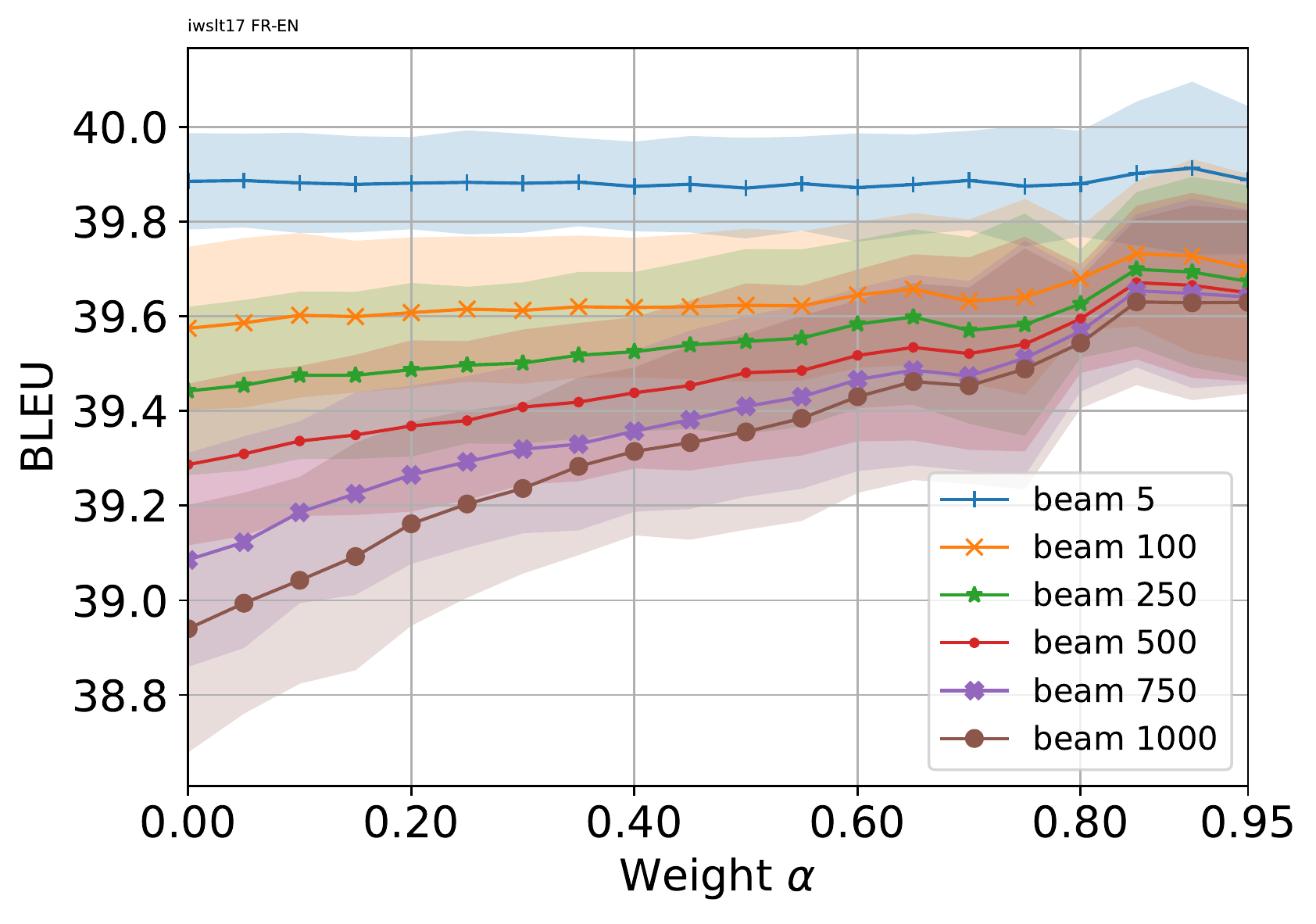}
         \caption{IWSLT'17 FR$\rightarrow$EN}
     \end{subfigure}
    \begin{subfigure}[b]{.49\linewidth}
         \centering
         \includegraphics[width=\linewidth]{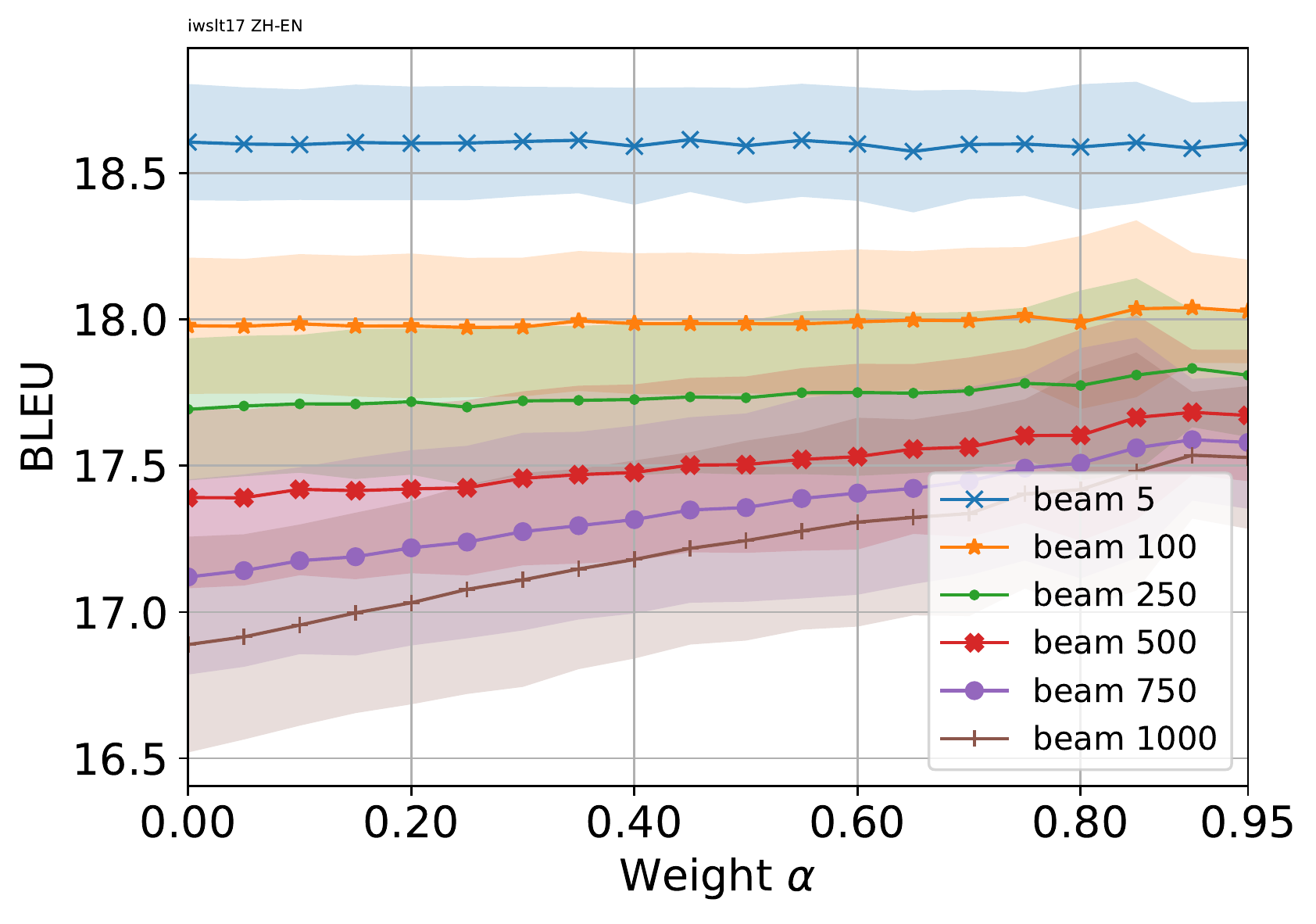}
         \caption{IWSLT'17 ZH$\rightarrow$DE}
     \end{subfigure} \\
    \begin{subfigure}[b]{.49\linewidth}
         \centering
         \includegraphics[width=\linewidth]{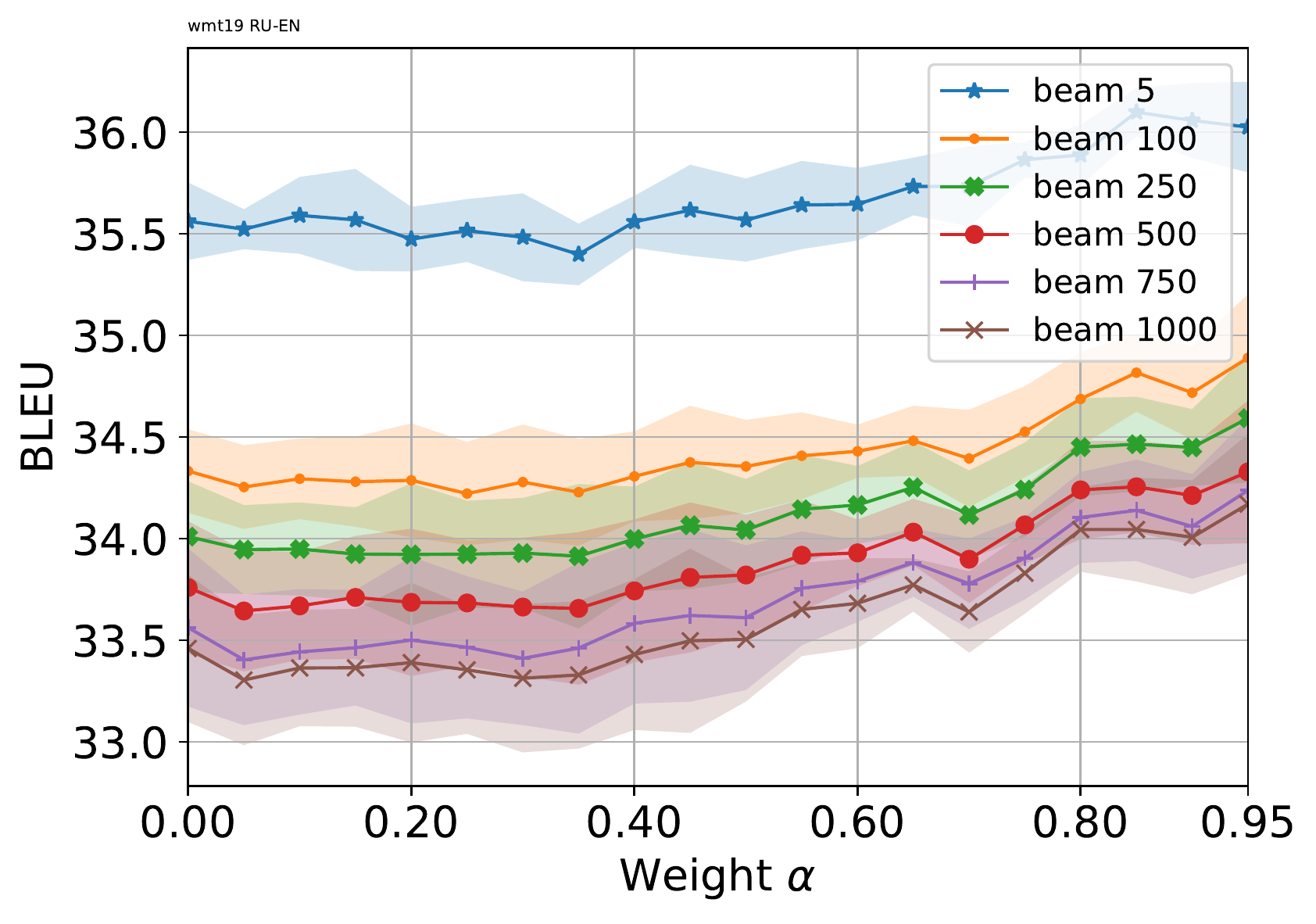}
         \caption{WMT'19 RU$\rightarrow$EN}
     \end{subfigure}
     \begin{subfigure}[b]{.49\linewidth}
         \centering
         \includegraphics[width=\linewidth]{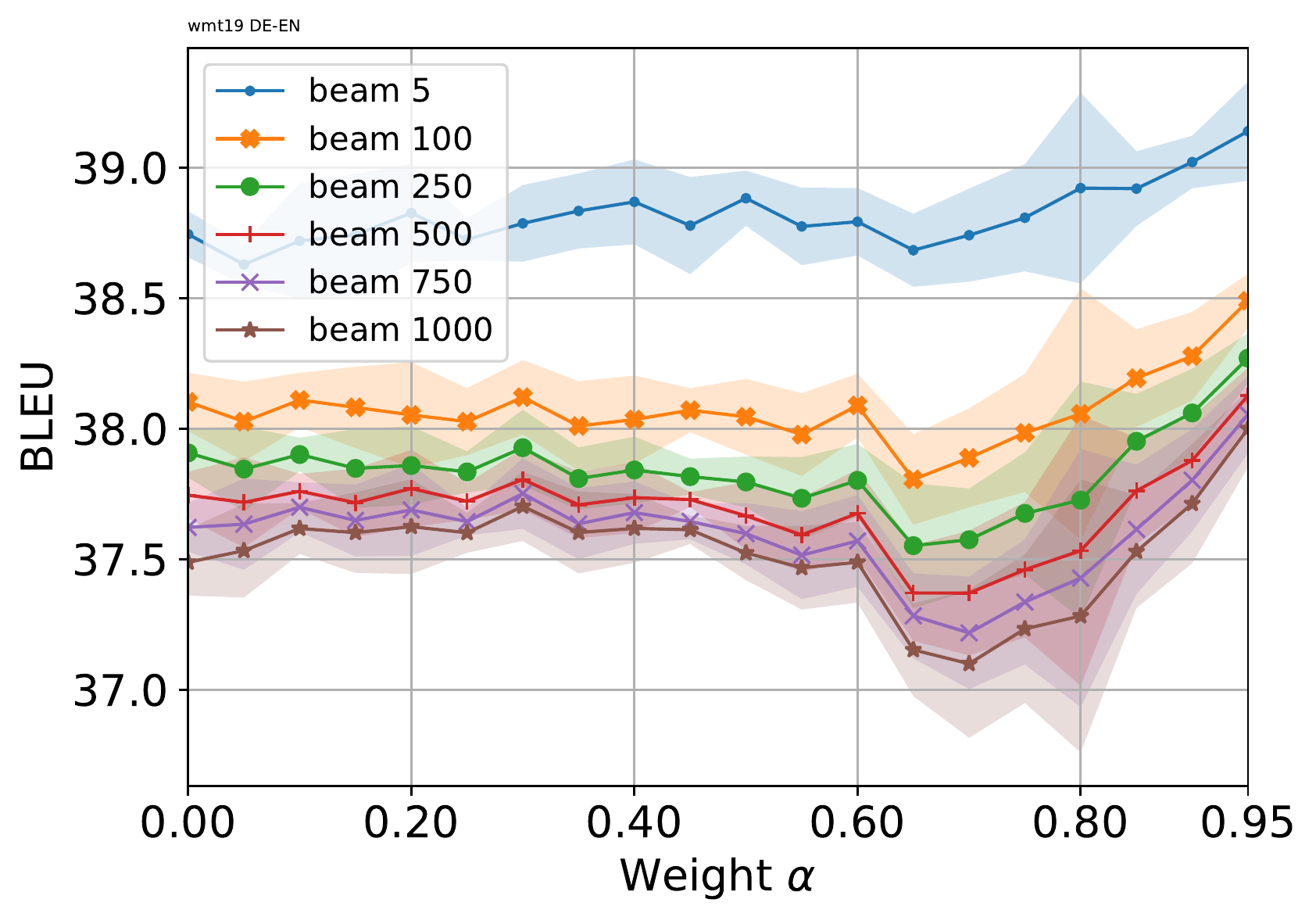}
         \caption{WMT'19 DE$\rightarrow$DE}
     \end{subfigure}
    \caption{BLEU score is measured on corresponding test sets. Decoding is done using beam search with beam sizes given in the legend. \cref{sec:exp_settings} provides more details on test sets and decoding hyper-parameters. Filled regions denote the standard deviation across training runs according to \cref{sec:exp_settings}.}
    \label{fig:bleu_four_systems_appendix}
\end{figure*}

\cref{fig:bleu_four_systems_appendix} presents BLEU scores measured on the rest of models which are not reported in the main text. We observe similar trends according to the discussion of \cref{fig:bleu_three_systems} in \cref{sec:experiments}.

\end{document}